\documentclass[lettersize,journal]{IEEEtran}
\usepackage{amsmath,amsfonts}
\usepackage{algorithmic}
\usepackage{algorithm}
\usepackage{array}
\usepackage[caption=false,font=normalsize,labelfont=sf,textfont=sf]{subfig}
\usepackage{textcomp}
\usepackage{stfloats}
\usepackage{url}
\usepackage{verbatim}
\usepackage{graphicx}
\usepackage{cite}
\usepackage{cuted}
\usepackage{svg}
\usepackage{booktabs}
\usepackage{multirow}
\usepackage{array}
\usepackage{tabularx}
\usepackage{amsmath}
\usepackage[acronym]{glossaries}
\makeglossaries
\newcommand{\alphabeta}[0]{{\alpha\beta}}

\newcommand{\kT}[1]{{ \ifnum#1>0 {(k+1)T_s} \else {kT_s} \fi}}

\begin{document}

\title{From Continuous Design to Delay-Aware Discrete Synthesis: Guaranteed High-Bandwidth Joint Control for PMSM Drives}

\author{Edmundo Pozo Fortuni\'c$^{1,*}$,~\IEEEmembership{Student Member,~IEEE,} Mehmet C. Yildirim$^{1}$,~\IEEEmembership{Member,~IEEE,}\\ Sami Haddadin$^{2}$, ~\IEEEmembership{Fellow,~IEEE}
        % <-this % stops a space
\thanks{$^{1}$The authors are with the Chair of Robotics and Systems Intelligence and the Munich Institute of Robotics and Machine Intelligence (MIRMI), Technical University of Munich, Germany. $^{*}$Corresponding author: \tt\small{edmundo.pozo@tum.de}}% <-this % stops a space
\thanks{$^{2}$Sami Haddadin is with Mohamed Bin Zayed University of Artificial Intelligence, Abu Dhabi, United Arab Emirates.} 
\thanks{This work was partially supported by the funding of the European Union's Horizon 2020 research and innovation programme as part of the project DARKO under grant no. 101017274.}
%\thanks{Manuscript received April 19, 2021; revised August 16, 2021.}
}

%\markboth{IEEE/ASME Transaction on Mechatronics}%
\markboth{Preprint submitted to arXiv}{Pozo Fortunić \MakeLowercase{\textit{et al.}}: Control Design and Synthesis for Discrete-Time PMSM Current Controllers of Robotic Joints}

\maketitle

\begin{abstract}
The increasing dynamic demands of modern robotic joints require current controllers to achieve high bandwidth over wide operating ranges of speed, acceleration, and torque, where communication, computation, and discrete-time effects can no longer be neglected. Conventional PMSM current controllers are typically designed in continuous time and subsequently discretized, leaving the sampling frequency and the impact of implementation delays largely to heuristic selection and iterative validation. This paper introduces a task-aware, delay-extended discrete-time joint model that explicitly accounts for physical communication and computation delays and enables direct synthesis of a discrete PI current controller with prescribed bandwidth and delay guarantees throughout the operating envelope. The framework analytically determines the minimum required sampling frequency, controller gains, and DC-link voltage needed to satisfy the specified motor and joint performance. Simulations across a range of dynamic requirements validate the methodology and demonstrate substantially reduced sampling-frequency and DC-link-voltage requirements compared with conventional continuous-time-based design. Experiments on a newly developed custom robotic joint further validate the proposed framework under real embedded implementation conditions.\end{abstract}


\begin{figure}[!t]
    \centering
     \includegraphics[width=0.75\linewidth]{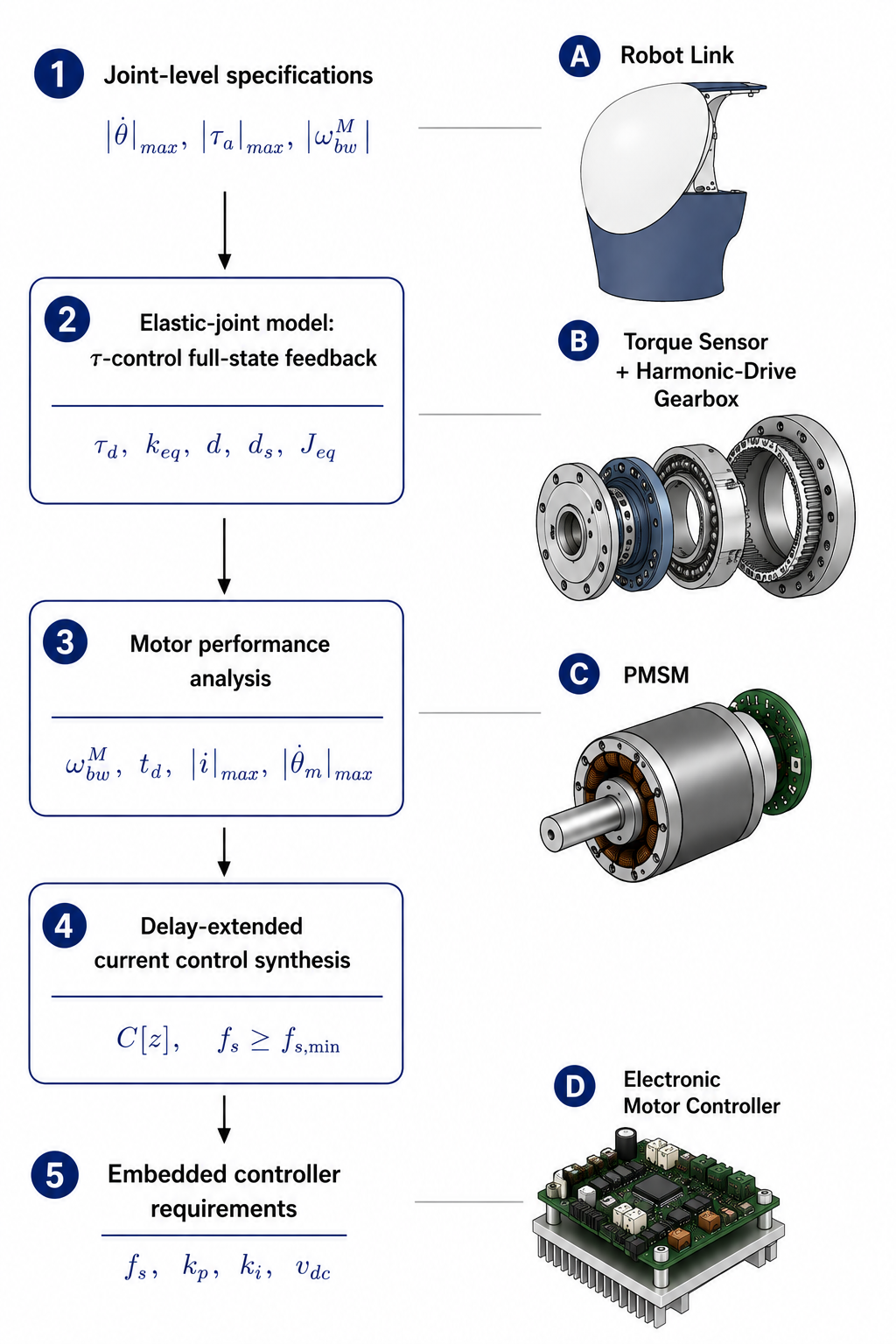}
    \caption{Design Synthesis for a PMSM embedded discrete controller based on its joint-level specifications.}
    \label{fig:enter-label}
\end{figure}

\section{Introduction}
 \IEEEPARstart{P}{ermanent}-magnet synchronous motors (PMSMs) are widely employed in modern robotic actuators due to their high torque density, efficiency, and dynamic capabilities \cite{hematiRobustNonlinearControl1990}. Their current control is commonly based on field-oriented control (FOC), where the motor dynamics are transformed into the rotating (dq) reference frame to independently regulate the torque-producing current \cite{blaschkePrincipleFieldOrientation1972}. With the introduction of digital implementations, early approaches retained the continuous-time (CT) design while assuming sufficiently high sampling frequencies to neglect discrete effects \cite{gabrielFieldOrientedControl1980}. Subsequent works introduced the use of complex-vector state variables and PI current control to simplify the system equations  \cite{rowanNewSynchronousCurrent1986,brizAnalysisDesignCurrent2000} while others introduced some discretization delays into their controller design through equivalent continuous representations \cite{holtzDesignFastRobust2004,tiapkinCurrentControllerDesign2020}.

Initial approaches to DT modeling relied on first-order Euler approximations of the continuous (dq)-model \cite{mohamedDesignImplementationRobust2007}. More accurate DT models were later developed to explicitly account for discretization effects under different sampling and PWM strategies \cite{kimDiscreteTimeCurrentRegulator2010,hoffmann2015digital,busada2018comments}, establishing a proper discretization in the rotating ($dq$) reference frame. While these works primarily addressed grid-connected converter applications, the DT model was later extended to PMSMs by incorporating the back-EMF \cite{Zhang2021deadbeat}. Despite these developments, simplified DT models based on first-order approximations remain in use \cite{walzDahlinBasedFast2019,smidt2021discrete}, typically assuming a sufficiently high sampling frequency for the neglected discretization effects to remain insignificant.

Several DT current-control approaches have also been proposed. Conventional PI regulators remain widely adopted \cite{kimDiscreteTimeCurrentRegulator2010,hoffmann2015digital,busada2018comments}, while nonlinear methods such as sliding-mode control \cite{smidt2021discrete} and observer-based approaches for compensating model uncertainties \cite{mohamedDesignImplementationRobust2007} have also been explored. Deadbeat control has additionally been proposed together with predictor-based compensation of the inherent digital delay \cite{Zhang2021deadbeat}.

The performance analysis of these approaches has focused on different aspects of the current-control problem. Some works emphasize the decoupling achieved by the controller architecture \cite{hoffmann2015digital,busada2018comments}, while others focus on robustness against parameter variations and model uncertainties \cite{mohamedDesignImplementationRobust2007,smidt2021discrete}. Comparisons between CT-based and DT approaches have also shown a degradation in the performance of CT-based designs \cite{kimDiscreteTimeCurrentRegulator2010}, although the DT effects responsible for this degradation are not explicitly related to the resulting closed-loop response. A more direct relation to performance is established by deadbeat approaches. In \cite{busadaSynchronousReferenceFrame2020}, the PI controller is parametrized to achieve a two-sample deadbeat reference response, while other works analyze the limits between tracking, disturbance rejection, and robustness \cite{Liao2017deadbeat}. Predictor-based approaches alternatively seek to compensate the inherent digital delay through model-based prediction \cite{Zhang2021deadbeat}. Nevertheless, these approaches generally target specific closed-loop properties rather than providing a systematic synthesis from arbitrary prescribed performance requirements.

Defining this relationship becomes particularly relevant in robotic actuators, where the required motor performance ultimately originates from joint-level requirements. In lightweight and intrinsically elastic robots, PMSMs are commonly coupled to harmonic-drive gearboxes\cite{iskandarJointLevelControl2020} and elastic elements, introducing dynamics between the electromagnetic motor torque and the actuator output torque \cite{Spong1987}. These dynamics were systematically incorporated into joint-control design in \cite{UnifiedPassivitybasedControl2007}, reducing the reliance on heuristic design choices while considering the underlying motor actuation as ideal. As operating demands increase, however, the performance required from the controlled motor must also be established to ensure the behavior prescribed by the outer control loops. This provides a performance-based abstraction between the joint- and motor-level dynamics, avoiding the need to incorporate the complete motor dynamics into the joint-level design while ensuring that their influence remains within prescribed limits. Establishing this relationship further enables the embedded motor-controller requirements to be derived from the actuator performance. In particular, the sampling frequency represents a limited computational resource, while the achievable current dynamics are constrained by the available dc-link voltage; both can therefore be treated as hardware requirements derived from the prescribed motor performance.

This paper presents a discrete-time design and synthesis framework for PMSM current controllers of robotic joints, linking joint-level performance requirements to the motor-control and embedded-system capabilities required to achieve them. From a prescribed closed-loop bandwidth and maximum admissible equivalent delay, together with the motor operating envelope, the framework determines the PI controller gains, minimum sampling frequency, and required dc-link voltage. Section II establishes the target motor performance and its relation to flexible-joint control and digital implementation. Section III presents the motor model and controller synthesis, including the determination of the controller gains, sampling frequency, and dc-link voltage. Section IV derives the conditions under which conventional continuous-time design can approximate the resulting response. The framework is evaluated in Section V, followed by a discussion in Section VI and conclusions in Section VII.

\section{Problem Statement}

\subsection{PMSM Continuous Model}
The continuous-time model of a surface-mounted permanent-magnet synchronous motor (PMSM) is well established. In the stationary $abc$-reference frame, the electrical dynamics are given by
\begin{equation} 
\label{eq:i_abc} 
\frac{d}{dt}{{i}}_{abc} =-\frac{R}{L}{i}_{abc} +\frac{1}{L} \left({v}_{abc}-{\epsilon}_{abc}\right), 
\end{equation} 
where $i$, $v$, and $\epsilon$ denote current, voltage, and back-EMF, respectively, with the subscript identifying their reference frame; $R$ and $L$ denote the phase resistance and inductance. For a sinusoidal back-EMF, the back-EMF vector and electromagnetic torque, $\tau_m$, are
defined as
\begin{equation}
\label{eq:emf_abc}
%\begin{split}
{\epsilon}_{abc}=k_t\dot{\theta}_e
\begin{bmatrix}
-\sin(\theta_e)\\
-\sin(\theta_e-2\pi/3)\\
-\sin(\theta_e+2\pi/3)
\end{bmatrix}\!\!, \quad
\tau_m=\dot{\theta}_e^{-1}{\epsilon}_{abc}^{\top}{i}_{abc}.
%\end{split}
\end{equation}
Where, $k_t$ is the torque/back-emf constant, $\theta_e$ and $\dot{\theta}_e$ are the electrical rotor position and angular velocity. Their relationship with the rotor mechanical position, $\theta_r$, is set by the number of pole pairs, $n_p$, and defined by 
\begin{equation}
    \theta_e = n_p\theta_r, \quad \dot{\theta}_e = n_p \dot\theta_r.
\end{equation}

Applying the Clarke transformation \cite{Gabriel1980FieldOrientedCO}  to (\ref{eq:i_abc}) and (\ref{eq:emf_abc}) maps the three-phase quantities into the stationary $\alpha\beta$-reference frame, where complex vectors are denoted in bold, and yields
\begin{equation}
\label{eq:i_alphabeta}
\!\frac{d}{dt}{\boldsymbol{i_{\alpha\beta}}}=-\frac{R}{L}\boldsymbol{i_{\alpha\beta}}+\frac{1}{L}\left(\boldsymbol{v_{\alpha\beta}}-\boldsymbol{\epsilon_{\alpha\beta}}\right)\!, \quad \boldsymbol{\epsilon_{\alpha\beta}}=jk_t\dot{\theta}_e e^{j\theta_e}\!.\!
\end{equation}
The Park transformation then maps a complex vector $\boldsymbol{x}$ from the $\alpha\beta$ frame into the synchronous $dq$-frame according to
\begin{equation}
\label{eq:X_dq}
\boldsymbol{x_{\alpha\beta}}=\boldsymbol{x_{dq}}e^{j\theta_e}, \quad
{\boldsymbol{ \dot x_{\alpha\beta}}}=\left({\boldsymbol{\dot x_{dq}}}+j\dot{\theta}_e\boldsymbol{x_{dq}}\right)e^{j\theta_e}.
\end{equation}
Applying (\ref{eq:X_dq}) to (\ref{eq:i_alphabeta}) gives
\begin{equation}
\label{eq:i_dq}
\!\frac{d}{dt}{\boldsymbol{i_{dq}}}=-\!\!\left(\!\!\frac{R}{L}+j\dot{\theta}_e\!\!\right)\!\boldsymbol{i_{dq}}+\frac{1}{L}\!\!\left(\boldsymbol{v_{dq}}-\boldsymbol{\epsilon_{dq}}\right)\!, \!
\quad \boldsymbol{\epsilon_{dq}}=jk_t\dot{\theta}_e. \!\!
\end{equation}
The electromagnetic torque then reduces to
\begin{equation}
    \tau_m = \frac{3}{2}k_t\operatorname{Im}\{\boldsymbol{i_{dq}}\}
\end{equation}
Thus, torque can be directly regulated through the quadrature component of the current vector.
\subsection{Continuous Current/Torque Control}

The control voltage is defined then by two terms,
\begin{equation}
\label{eq:vt}
    \boldsymbol{v_t}= \boldsymbol{v_c}+\boldsymbol{v_\mathrm{ff}}
\end{equation}
where $\boldsymbol{v_t}$, $\boldsymbol{v_c}$, and $\boldsymbol{v_\mathrm{ff}}$ denote the total, feedback-control and feedforward voltages, respectively. The feedforward term compensates the speed-dependant cross-coupling and back-EMF terms of the dq-frame model as, 
\begin{equation}
\boldsymbol{v_\mathrm{ff}}= jL{\dot \theta_e}\boldsymbol{i_{dq}} + \boldsymbol{\epsilon_{dq}}  
\end{equation}
Under ideal compensation, the current dynamics reduce to the first-order system
\begin{equation}
    H(s) = \frac{I_{dq}(s)}{V_c(s)}=\frac{1}{L}\frac{1}{s+\frac{R}{L}}
\end{equation}
The feedback control voltage is generated using a PI controller,
\begin{equation}
    C(s) = \frac{K_{p,i}}{s}( s + T_{i,i})
\end{equation}
where $K_{p,i}$ is the proportional gain and $T_{i,i}$ defines the controller-zero frequency (together they form the integral gain). By placing the controller zero at the electrical pole and selecting the
proportional gain according to the desired bandwidth,
$T_{i,i}=R/L$ and $K_{p,i}=L\omega_{bw}$, respectively, the plant pole
is cancelled. The resulting closed-loop transfer function describes the
current response from the commanded current $\boldsymbol{i}_{dq}^{*}$ to
$\boldsymbol{i}_{dq}$ and, equivalently, the torque response from the
commanded torque $\tau_{\mathrm{cmd}}$ to $\tau_m$, a
\begin{equation}
\label{eq:hcl}
    H_{m}(s) = \frac{I_{dq}(s)}{I^*_{dq}(s)}= \frac{\tau_m(s)}{\tau_\mathrm{cmd}(s)}=\frac{\omega_{bw}}{s+\omega_{bw}}
\end{equation}
Thus, under ideal continuous-time compensation, both the current and
torque responses follow the same first-order dynamics, with zero
steady-state error and a prescribed bandwidth $\omega_{bw}$ directly
determined by $K_{p,i}$.
% \begin{figure}[!t]
%     \centering
%     \includesvg[width=\linewidth]{figures/PhD_BLDC_controller-Robot Controller_white.svg}
%     \caption{replace figure with something simpler}
%     \label{fig:sea_ctrl}
% \end{figure}
\subsection{Inherent Delay in Digital Implementation}

The closed-loop response in (\ref{eq:hcl}) represents the ideal performance achievable under a fully continuous-time implementation. In practice, however, the motor controller is digitally implemented and must therefore adhere to the discrete-time nature of signal acquisition, computation, and voltage actuation. Consequently, even if the digital controller perfectly reproduces the target first-order dynamics, its response cannot be equivalent to (\ref{eq:hcl}) without accounting for the delay inherently introduced by the digital implementation.

In the ideal case, the digital implementation introduces one sampling period $T_s$ of computational delay, while the zero-order hold (ZOH) associated with the voltage actuation contributes an additional equivalent delay of $T_s/2$. \cite{} Therefore, even under otherwise ideal conditions, the equivalent delay $T_d$ is bounded.. Accordingly, the best achievable equivalent representation of the controlled motor is
\begin{equation}
\label{eq:Hm_simplified}
H_m^s(s)=\frac{\omega_{bw}}{s+\omega_{bw}}e^{-sT_d}, \qquad T_d \geq 1.5T_s,
\end{equation}

Hence, $T_d=1.5T_s$ constitutes the ideal lower bound rather than an additional design choice, while other discrete-time effects may further increase the equivalent delay.

\section{Discrete Controller Synthesis}

A key consequence of the discrete-time formulation is that the Park transformation in (\ref{eq:X_dq}) is evaluated only at the sampling instants, becoming piecewise constant over each sampling interval,

\begin{equation}
\begin{split}    
\boldsymbol{x_{\alphabeta}}(t) =& \boldsymbol{x_{dq}}(t) \!\cdot\! e^{j\theta_e[kT_s]}, \quad \quad t\in [kT_s,(k+1)T_s[ \\
 &\boldsymbol{x_{dq}}(t) \!\cdot\! e^{j\theta_e[(k\!+\!1)T_s]}, t\in [(k\!+\!1)T_s,(k\!+\!2)T_s[
\end{split}
\end{equation}

where, assuming constant electrical angular velocity within each sampling interval,
\begin{equation}
\theta_e[k+1]=\theta_e[k]+\Delta_e[k],\qquad \Delta_e[k]=T_s\dot{\theta}_e[k].
\end{equation}

Therefore, the stationary $\alpha\beta$ model is first discretized and subsequently transformed into the $dq$ frame. This ordering preserves the electrical-angle evolution over each sampling interval and introduces the corresponding discrete rotation into the model.

\subsection{Discrete PMSM Model}
Applying exact discretization to the stationary-frame current dynamics yields
\begin{equation}
\begin{split}
\boldsymbol{i_{\alpha\beta}}((k+1)T_s) = & e^{-\frac{R}{L}T_s}\boldsymbol{i_\alphabeta}(kT_s)\\
+\frac{1}{L}\int_{k}^{(k+1)T_s}&e^{-\frac{R}{L}((k+1)T_s-t)}(\boldsymbol{v_\alphabeta}(t)-\boldsymbol{\epsilon_\alphabeta}(t))dt
\end{split}
\end{equation}
Assuming also the applied voltage and the electrical speed remain constant within each sampling interval, the exact discrete-time solution becomes

\begin{equation}
\begin{split}  
\boldsymbol{i_\alphabeta}[k+1] &= \underbrace{e^{-\frac{R}{Lf_s}}}_{:=a}\boldsymbol{i_\alphabeta}[k]+\underbrace{\frac{1-e^{-\frac{R}{Lf_s}}}{R}}_{:=b}\boldsymbol{v_\alphabeta}[k]\\
&-\frac{k_t}{L} e^{j\theta_e[k]}(1-e^{-\frac{R}{Lf_s}}e^{-j\frac{n_p \dot \theta}{f_s}}) \frac{jn_p \dot \theta}{\frac{R}{L}+jn_p \dot \theta} 
\end{split}
\end{equation}
Evaluating the discrete Park transformation at the next sampling instant gives
\begin{equation}
  \boldsymbol{x_{\alphabeta}}[k+1] = \boldsymbol{x_{dq}}[k+1] \cdot  e^{j\theta_e[k]}\cdot \underbrace{e^{jT_s\dot \theta_e[k]}}_{\boldsymbol{r_\omega^{-1}} }
\end{equation}
Substituting this relation into the discrete $\alpha\beta$-frame model results in the following discrete motor model expressed in the rotating dq-reference frame:
\begin{equation}
\begin{split}  
\boldsymbol{i_{dq}}[k+1] &=\underbrace{a\boldsymbol{r_\omega}}_{:=\boldsymbol{a_\omega}}\boldsymbol{i_{dq}}[k]+\underbrace{b\boldsymbol{r_\omega}}_{:=\boldsymbol{b_\omega}} ( \boldsymbol{v_{dq}}[k]-\boldsymbol{\epsilon_{dq}}[k])\\
\boldsymbol{\epsilon_{dq}}[k]=&k_t\underbrace{\frac{\boldsymbol{r_\omega}-a}{1-a}}_{:=\boldsymbol{c_\omega}} \frac{j\frac{R}{L} \dot \theta_e}{\frac{R}{L}+j\dot \theta_e} )
\end{split}
\end{equation}
The zero-order-hold (ZOH) behavior introduced by the digital controller can be equivalently represented as a one-sample delay in the plant model \cite{}. Incorporating this delay yields the final discrete-time model

\begin{equation}
\begin{split}  
\boldsymbol{i_{dq}}[k+1] &=\boldsymbol{a_{\omega}}\boldsymbol{i_{dq}}[k]+\boldsymbol{b_\omega}(\boldsymbol{ r_\omega}\boldsymbol{v_{dq}}[k-1]-\boldsymbol{\epsilon_{dq}}[k])\\
\end{split}
\end{equation}

\subsection{Step 1: Controller Architecture and Gains}

\begin{figure}[t]
    \centering
     %\includesvg[width=\linewidth]{figures/kp_bw_delay.svg}
     \includegraphics[width=\columnwidth]{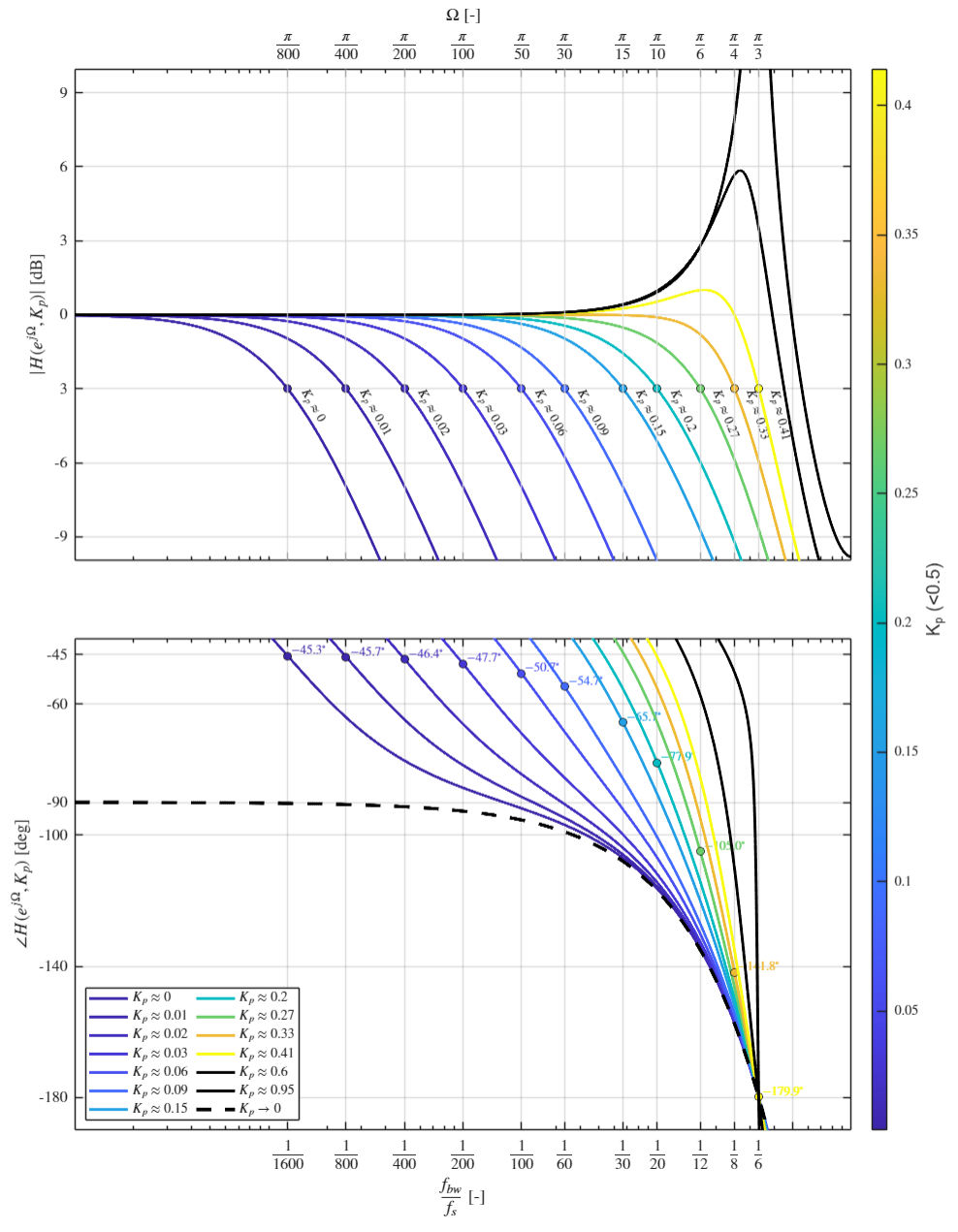}
    \caption{Magnitudes (top) and Phases (bottom) of the closed loop current controllers \ref{eq_hcl_i} with different $Kp$ parameters. Each $\Omega_bw$ is highlighted and marked on the top axis. The bottom axis has the equivalent $f_{bw}/f_s$  ratio. Each corresponding phase is also marked.}
    \label{fig:kp}
\end{figure}

To compensate for the back electromotive force (back-EMF), the following feedforward voltage is introduced:
\begin{equation}
\label{eq:vff_d}
    \boldsymbol{v^d_\mathrm{ff}}[k]= \boldsymbol{r_\omega^{-1}\epsilon_{dq}}[k]
\end{equation}
yielding the discrete complex plant
\begin{equation}
H^d(z)=\frac{\boldsymbol{b_\omega r_\omega}}{z(z-\boldsymbol{a_w})}
\end{equation}
Selecting the discrete complex PI controller
\begin{equation}
    C^d[z]= \frac{K_{p,i}}{\boldsymbol{b_\omega r_\omega}}\frac{(z-\boldsymbol{a_{i,i})}}{z-1}
\end{equation}
where $K_{p,i}$ and $a_{i,i}$ are the complex proportional and integral gains of the controller, defined by: 
\begin{equation}
    K_{p,i} = \frac{K^d_p}{\boldsymbol{b_\omega r_\omega}}
\end{equation}
\begin{equation}
a_{i,i} = e^{-\frac{R}{L}T_s}    
\end{equation}
the results in a second order closed-loop transfer function
\begin{equation}
\label{eq_hcl_i}
    H^d_{cl}(z,K_{p,i}) = \frac{K_{p,i}}{z^2-z+K_{p,i}}
\end{equation}
From the structure of (\ref{eq_hcl_i}) and the Routh--Hurwitz criterion, the system is stable for $0<K_{p,i}<1$. However, its closed-loop behavior varies significantly within this range. For $K_{p,i}\leq1/4$, the poles are real, resulting in a monotonic response, whereas for $1/4<K_{p,i}\leq1/3$, the poles become complex without introducing a resonant peak. The resulting responses for different values of $K_{p,i}$ are illustrated in Fig.\ref{fig:kp}. Therefore, before determining the gain associated with the target bandwidth, a preliminary upper bound $K_{p,i}^{\mathrm{max}}$ is selected according to the desired closed-loop behavior. The gain required to achieve the target bandwidth is then obtained by imposing the $-3$-dB condition

\begin{equation}
\label{eq:3db_defined}
|H^d_{cl}(e^{j\Omega_{bw}},K_{p,i})|^2=\frac{1}{2}.
\end{equation}

Here,

\begin{equation}
\Omega_{bw}=\omega_{bw}T_s
\end{equation}

denotes the normalized angular bandwidth. Solving \eqref{eq:3db_defined} for $K_{p,i}$ yields

\begin{equation}
\begin{split}
K_{p,i}^{bw}\!&=\cos2\Omega_{bw}-\cos\Omega_{bw} \\
&+\!\sqrt{2\!\left(\cos2\Omega_{bw}-\cos\Omega_{bw}\right)^2+\left(\sin2\Omega_{bw}-\sin\Omega_{bw}\right)^2}.
\end{split}
\end{equation}
The selection of $K_p^d$ determines not only the equivalent bandwidth but also the equivalent delay of the synthesized response, thereby coupling both quantities to $f_s$. At the closed-loop bandwidth, the ideal first-order target exhibits a phase lag of $-\pi/4$, whereas the phase of the synthesized response is given by

\begin{equation} \phi(\Omega,K_p)=\arctan\left(\frac{\sin 2\Omega-\sin\Omega}{\cos 2\Omega-\cos\Omega+K_p}\right). \end{equation}

The additional phase lag can therefore be represented by an equivalent delay $T_d$, such that

\begin{equation} \phi_{-3\mathrm{dB}}=-\frac{\pi}{4}-\omega_{\mathrm{bw}}T_d. \end{equation}

Since $K_p^{\mathrm{bw}}$ is determined by the normalized target bandwidth $\Omega_{\mathrm{bw}}=\omega_{\mathrm{bw}}T_s$, the resulting equivalent delay is also inherently coupled to the sampling frequency. This relationship is later used to establish a lower bound on $f_s$.

%[Preliminary elimination of Kp ranges beyond stability, eg Kp> sqrt(2)-1]

\subsection{Step 2: Sampling Frequency Boundaries}
\begin{figure*}[t]
    \centering
%     \includesvg[width=\linewidth]{figures/fs_calc}
     \includegraphics[width=\linewidth]{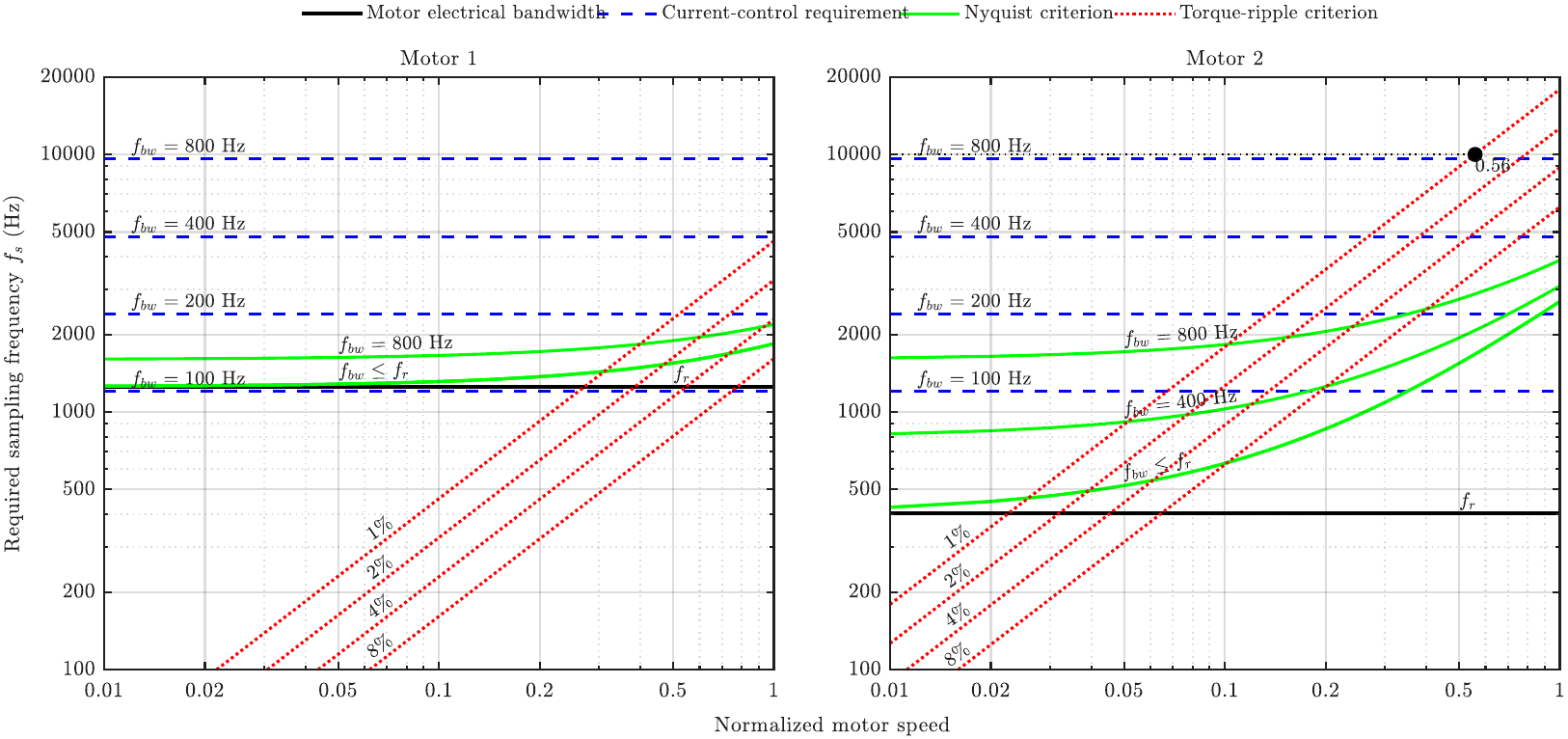}
    \caption{Boundaries required to select the correct minimum sampling frequency $f_s$ that ensure the proper operation of two motors (see Tab.\ref{tab:parameters}) for the required conditions of bandwidth, maximum speed and current ripple.}
    \label{fig:fs}
\end{figure*}
The minimum sampling frequency is constrained by three independent mechanisms: the acquisition of the physical phase currents, the discrete implementation of the rotating reference frame, and the closed-loop bandwidth and equivalent delay of the synthesized controller. Each condition establishes a lower bound on $f_s$ that must be simultaneously satisfied.

\subsubsection{Signal Acquisition}

Although the current controller is implemented in the synchronous discrete-time $(dq)$ reference frame, current acquisition is inherently performed in the stationary $(abc)$ frame, where the physical phase currents are sampled. Consequently, the sampling frequency must satisfy the Nyquist criterion with respect to the spectral content of the measured $(abc)$ currents rather than the transformed $(dq)$ quantities. Since the measured currents contain both the controller dynamics and the electrical rotation of the reference frame, the highest frequency component is given by the sum of the controller bandwidth and the electrical frequency. Therefore,

\begin{equation} f_s > 2\left(f_c+f_e\right), \end{equation}

where $f_e$ is the maximum electrical frequency and

\begin{equation} f_c=\max\left(f_r,f_{\mathrm{bw}}\right), \end{equation}

with $f_{\mathrm{bw}}$ denoting the desired closed-loop current bandwidth and

\begin{equation} f_r=\frac{R}{2\pi L}, \end{equation}

representing the electrical RL corner frequency of the motor.

\subsubsection{Discrete Park Transformation}

A second limitation originates from the discrete implementation of the Park transformation. Although the controller operates in the synchronous $(dq)$ frame, the transformation is updated only at the sampling instants, introducing a discrepancy between the continuously rotating current vector and its discrete representation over each sampling interval. This produces a periodic current error that manifests as torque ripple, as reported in~\cite{Yim2009Modified}.

By limiting the admissible torque-ripple ratio to $\gamma_q$, the sampling frequency must satisfy

\begin{equation} f_s > \frac{\omega_e}{2\arccos\left(1-2\gamma_q\right)}, \end{equation}

thereby ensuring that the degradation in torque production caused by the discrete Park transformation remains below the specified limit over the complete operating envelope.

\subsubsection{Closed-Loop Bandwidth and Equivalent Delay}

As previously established, the digital implementation imposes a theoretical lower bound of $T_d\geq1.5T_s$. To limit the additional delay introduced by the synthesized response, the proposed design imposes an upper bound of

\begin{equation} T_d\leq2T_s. \end{equation}

Using the phase relationship derived in the controller synthesis, this requirement imposes

\begin{equation} \phi(\Omega_{\mathrm{bw}},K_p^{\mathrm{bw}})\geq-\frac{\pi}{4}-2\Omega_{\mathrm{bw}}. \end{equation}

Substitution of $K_p^{\mathrm{bw}}$ yields

\begin{equation} 4\left(\cos\Omega_{\mathrm{bw}}-1\right)\left(\sin\Omega_{\mathrm{bw}}-\sin^2\frac{3\Omega_{\mathrm{bw}}}{2}\right)\leq0. \end{equation}

which is satisfied for

\begin{equation} 0<\Omega_{\mathrm{bw}}\leq\frac{\pi}{6}. \end{equation}

Therefore,

\begin{equation} f_s\geq\frac{6}{\pi}\omega_{\mathrm{bw}}=12f_{\mathrm{bw}}. \end{equation}

This establishes the sampling-frequency boundary required to achieve the prescribed bandwidth while maintaining the equivalent delay within the specified limit.

\subsubsection{Minimum Sampling Frequency}

The three previous constraints arise from independent mechanisms and must therefore be satisfied simultaneously. The minimum feasible sampling frequency is consequently given by

\begin{equation} f_s\geq \max\left\{2\left(f_c+f_e\right);\frac{\omega_e}{2\arccos\left(1-2\gamma_q\right)};12f_{\mathrm{bw}}\right\}. \end{equation}

This ensures aliasing-free current acquisition, bounds the torque ripple introduced by the discrete Park transformation, and provides the closed-loop bandwidth and equivalent-delay performance required from the synthesized current controller.

\subsection{Step 3: Required DC-link voltage}
\subsubsection{Nominal $dq$ Voltage}

Once the sampling frequency is selected according to the previous criteria, the corresponding controller gain $K_{p,i}^{\star}$ is fully determined. The voltage required to produce the commanded current can then be evaluated over the specified motor operating envelope. From the synthesized closed-loop response, the control-voltage transfer function is defined as

\begin{equation}
\begin{split}    
G(z,K_{p,i},\dot \theta)&=\frac{\boldsymbol{V_{dq}}(z)}{\boldsymbol{I^*_{dq}}(z)}=  \frac{H^d(z)}{H^d_{cl}(z)}  \\
&= \frac{K_{p,i}}{b_\omega r_\omega}\frac{z^2-a_\omega z}{z^2-z+K_{p,i}}
\end{split}
\end{equation}

For the selected gain and maximum operating speed, the maximum control-voltage gain is obtained as

\begin{equation}
\label{eq:Gmax}
{|G|}^\mathrm{max} = |G(e^{j\Omega_{G^{\mathrm{max}}}^{\star}},K^{\star}_{p,i},\dot \theta ^{\mathrm{max}})|
\end{equation}
where the corresponding frequency is determined from
\begin{equation}
\Omega_{G^{\mathrm{max} }}^{\star}=\operatorname*{arg\,solve}_{\Omega\in[0,\pi]}\!\left(\frac{d}{d\Omega}\left( \left|G\!\left(e^{j\Omega},K^{\star}_{p,i},\dot \theta^{\mathrm{max}} \right) \right| ^2\right)\!=0\! \right)
\end{equation}

The maximum voltage can be then calculated using (\ref{eq:vt}) with the feedforward voltage defined in \eqref{eq:vff_d} over the prescribed current $\boldsymbol{i_{dq}}^{\mathrm{max}}$ and speed range $\dot \theta^{\mathrm{max}}$ can then be obtained by:
\begin{equation}
\begin{split}   
\boldsymbol{u^\mathrm{max}}&= {|G|}^\mathrm{max} \cdot |\boldsymbol{i_{dq}}|^\mathrm{max}+|\boldsymbol{v_\mathrm{ff}}(\dot \theta ^\mathrm{max})|
\end{split}
\end{equation}

The resulting $\boldsymbol{u}^{\mathrm{max}}$ defines the voltage authority required over the prescribed motor operating envelope.

%we add a figure with the design synthesis.
\subsubsection{DC-Link Voltage}

The corresponding dc-link voltage depends on the selected modulation strategy. Since space-vector PWM (SVPWM) provides a higher utilization of the available dc-link voltage than sinusoidal PWM (SPWM), its minimum value is given by

\begin{equation}
V_{\mathrm{dc}}^{\mathrm{min}}=
\begin{cases}
2|\boldsymbol{u}^{\mathrm{max}}|, & \mathrm{SPWM},\\
\dfrac{2}{\sqrt{3}}|\boldsymbol{u}^{\mathrm{max}}|, & \mathrm{SVPWM}.
\end{cases}
\end{equation}

Together with $f_s$ and the controller gains, this completes the synthesis of the motor-controller requirements from the prescribed performance and operating envelope.

\begin{table}[t]
\centering
%\vspace*{0.3cm}
\caption{Motor parameters of the robotic joint prototype.}
\label{tab:parameters}
\begin{tabular}{|l|c|c|c|c|}
\hline
Parameter & Symbol & Unit & $M_1$ & $M_2$ \\ \hline \hline
Phase Resistance & $R$ & $\Omega$ & 12.65e-2 & 7.55e-2 \\ \hline
Phase Inductance & $L$ & $\mathrm{H}$ & 3.21e-5 & 5.95e-5 \\ \hline
Pole Pairs & $n_p$ & $-$ & 2 & 10 \\ \hline
\begin{tabular}[l]{@{}l@{}} Torque Constant \\ Back-EMF Constant \end{tabular}& $k_t$ &\begin{tabular}[l]{@{}l@{}} $\mathrm{N\cdot m/A}$ \\ $(\mathrm{V\cdot s/rad})$\end{tabular}& 2.81e-2 & 3.1e-2 \\ \hline 
Rated Current & $i_{\mathrm{max}}$ & $\mathrm{A}$ & 6.6 & 10 \\ \hline
Rated Speed & $\omega_{\mathrm{max}}$ & $\mathrm{rad/s}$ & 924 & 717 \\ \hline
\begin{tabular}[l]{@{}l@{}}Maximum Electrical \\ Frequency \end{tabular} & $f_{e,\mathrm{max}}$ & $\mathrm{Hz}$ & 294e0 & 1.14e3 \\ \hline
Rated Torque & $\tau_{\mathrm{max}}$ & $\mathrm{N\cdot m}$ & 158e-3 & 320 \\ \hline
Maximum Power & $P_{\mathrm{max}}$ & $\mathrm{W}$ & 152 & 215 \\ \hline
Rotor Inertia & $J_m$ & $\mathrm{kg\cdot m^2}$ & 2.38e-5 & 1.34e-5 \\ \hline
\end{tabular}
\vspace{-0.7cm}
\end{table}
%\subsection{Simulation}

%\subsection{Experiments}
%\subsubsection{Experimental Results}

%\section{Discussion}

%\subsection{Model under acceleration}

%\section{Conclusion}
%The conclusion goes here.

\section*{Acknowledgments}
The authors thank J. Ringwald, A. Rahmon, and A. Wiedermann for their valuable support.

{\appendices
\section{Motor requirements for Flexible Joint Performance}
Following Abu-Schaffer's et al \cite{UnifiedPassivitybasedControl2007} and subsequent developments on flexible-based joints \cite{}, we know that the performance of the torque is defined by the torque control shaping under the overlying application (e.g. Impedance Control and/or Force Control, Trajectory Generation, etc.). Nevertheless, all of this applications consider an ideal performance of the motor controller; which, in principle does not allow us to quantify the target performance. Figure \ref{fig:sea_ctrl} shows a simplified example of a Unified Force-Impedance Controller for a flexible-joint \cite{Haddadin2024Unified}. The inner blocks have been extended to show the equivalent controlled motor model (\ref{eq:Hm_simplified}) in series of the classical approach. This of course, would affect the performance of the calculated values followin the traditional criteria defined in \cite{UnifiedPassivitybasedControl2007}. Therefore, a clear analysis including the effects of (\ref{eq:Hm_simplified}) is presented.

\subsubsection{Flexible Joint Model}
As stated in \cite{Spong1987}, the full system model can be defined as
\begin{equation}
\label{eq:robot_dyna}
    M(q)\ddot q+C(q,\dot q)+g(q) = \tau_a+\tau_\mathrm{ext}
\end{equation}
\begin{equation}
\label{eq:tau_a}
   \tau_a = \tau_s+DK^{-1}\dot \tau_s
\end{equation}
\begin{equation}
\label{eq:tau_s}
   \tau_s = K(\theta-q)
\end{equation}
\begin{equation}
\label{eq:motor_dyna}
    B\ddot \theta + \tau_a = \tau_m-\tau_f
\end{equation}
where the vectors $q,\theta \in \mathbb{R}^n $ contain the link and motor output positions. In (\ref{eq:robot_dyna}), $M(q) \in \mathbb{R}^{n\times n}$,  $C(q,\dot q)$, $g(q)$ and $\tau_\mathrm{ext} \in \mathbb{R}^{n}$ are  the inertia matrix, the centripetal and Coriolis vector and the gravity vector, and the external torques acting on the robot,respectively. Additionally, $\tau_a, \tau_s \in \mathbb{R}^{n}$ are the total actuator torque propagated from the motor to the robot via the spring-damp system, and the spring torque itself. Its relation is defined in (\ref{eq:tau_a}), where $K=\mathrm{diag}(k_i)$ and $ D=\mathrm{diag}(d_i) \in \mathbb{R}^{n\times n}$ are the diagonal positive definite joint stiffness matrix and the diagonal positive semi-definite joint damping matrix respectively. Within the motor,  $ B=\mathrm{diag}(b_i) \in \mathbb{R}^{n\times n}$ represents the positive definite motor inertia matrix,and $\tau_m$ and $\tau_f$ are the motor electromagnetic torque and the friction torques, respectively.

\begin{table*}[t]
\caption{Equivalent actuator models and resulting closed-loop transfer functions.}
\label{tab:model_cases}
\centering
\setlength{\tabcolsep}{3pt}
\begin{tabular}{|
>{\centering\arraybackslash}p{1.4cm}|
c|
>{\centering\arraybackslash}p{1.25cm}|
>{\centering\arraybackslash}p{2.5cm}|
c|}
\hline

$\begin{gathered}\textbf{Actuator}\\\textbf{Model}\end{gathered}$ &
\textbf{Closed-Loop Transfer Function $\boldsymbol{H^{A,t}_{cl}(s)}$} &
$\boldsymbol{\omega_n^2}$ &
$\boldsymbol{\zeta}$ &
\textbf{Additional pole(s)} \\
\hline

$\begin{gathered}
\text{Ideal}\\[1mm]
1
\end{gathered}$ &
$\dfrac{1}{b_\theta}
\dfrac{k+ds}{s^2+2\zeta\omega_n s+\omega_n^2}$ &
$\dfrac{k}{J_\theta}$ &
$\dfrac{d_\theta}{2\omega_n}$ &
-- \\
\hline

$\begin{gathered}
\text{Delay}\\[1mm]
e^{-sT_d}
\end{gathered}$ &
$\dfrac{\frac{T_d}{2}s+1}{\frac1ps+1}
\!\left(\dfrac{2}{T_dp\,b_\theta}\right)
\dfrac{(k+ds)e^{-sT_d}}
{s^2+2\zeta\omega_n s+\omega_n^2}$ &
$\dfrac{2}{T_dp}\dfrac{k}{J_\theta}$ &
$\dfrac{2\frac{d}{J}-d_\theta+\frac{2}{T_d}-p}{2\omega_n}$ &
$p=\dfrac2{T_d}
-2\!\left(d_\theta-\dfrac dJ\right)
+\mathcal{O}(T_d)$ \\
\hline

$\begin{gathered}
\text{Finite}\\[1mm]
\dfrac{\omega_{bw}}{s+\omega_{bw}}
\end{gathered}$ &
$\dfrac{\omega_0}{s+\omega_0}
\!\left(\frac{\omega_{bw}}{\omega_0 b_\theta}\right)
\dfrac{k+ds}
{s^2+2\zeta\omega_n s+\omega_n^2}$ &
$\dfrac{\omega_{bw}}{\omega_0}\dfrac{k}{J_\theta}$ &
$\dfrac{\frac{d}{J}+\omega_{bw}-\omega_0}{2\omega_n}$ &
$\omega_0=\omega_{bw}
-\left(d_\theta-\dfrac dJ\right)
+\mathcal{O}(\omega_{bw}^{-1})$ \\ [5mm]
\hline

$\begin{gathered}
\text{Finite +}\\
\text{Delay}\\[1mm]
\dfrac{\omega_{bw}e^{-sT_d}}{s+\omega_{bw}}
\end{gathered}$ &
$\dfrac{\omega_0}{s+\omega_0}
\dfrac{\frac{T_d}{2}s+1}{\frac1ps+1}
\!\left(\dfrac{2\omega_{bw}}{T_dp\,\omega_0b_\theta}\right)
\dfrac{(k+ds)e^{-sT_d}}
{s^2+2\zeta\omega_n s+\omega_n^2}$ &
$\dfrac{2\omega_{bw}}{T_dp\omega_0}\dfrac{k}{J_\theta}$ &
$\!\!\dfrac{\frac{d}{J}\!+\!\frac{2}{T_d}\!\!-\!p\!+\!\omega_{bw}\!-\!\omega_0\!}{2\omega_n}$ &
$\begin{gathered}
p=\dfrac2{T_d}-\!\ \frac{\omega_{bw}\left(2d_\theta -\dfrac dJ\right)}{\frac{2}{T_d}-\omega_{bw}}+\mathcal{O}\left(\!T_d,\omega_{bw}^{-1}\!\right)\!\\[1mm]
\!\omega_0=\omega_{bw}\!-\! \frac{\frac{2}{T_d}\!\left(\!d_\theta\!-\!\dfrac dJ \!\right)\!+\! \omega_{bw}d_\theta}{\frac{2}{T_d}\!-\!\omega_{bw}}\!+\!\mathcal{O}\!\left(\!T_d,\omega_{bw}^{-1}\!\right)\!
\end{gathered}$ \\ [8mm]
\hline
\end{tabular}
\end{table*}

By extracting one of the joints of the system and suppressing the external, friction, gravity, and Coriolis forces, the actuator transfer function can be defined as
\begin{equation}
    \label{eq:H_sea}
    H^A(s)=\frac{\tau_a(s)}{\tau_m(s)}
    =\frac{1}{b}\frac{k+ds}{s^2+\frac{d}{J}s+\frac{k}{J}}.
\end{equation}
where $J$ is the equivalent joint inertia, defined by
\begin{equation}
    J^{-1}=b^{-1}+m^{-1}.
\end{equation}
The denominator in (\ref{eq:H_sea}) defines a second-order mode characterized by its natural frequency $\omega_n$ and damping ratio $\zeta$, while its steady-state torque ratio is given by
\begin{equation}
    \label{eq:sea_modal}
    \omega_n=\sqrt{\frac{k}{J}},\qquad
    \zeta=\frac{d}{2\sqrt{kJ}},\qquad
    \left.\frac{\tau_a}{\tau_m}\right|_{s=0}=\frac{J}{b}.
\end{equation}
These quantities characterize the dominant joint dynamics and will be used to evaluate how the motor-control dynamics modify the resulting joint response.

\subsubsection{Joint Torque Controller}
As defined by \cite{UnifiedPassivitybasedControl2007},the commanded torque is assumed to be equivalent to the motor torque, thus the controller takes the shape defined as:
\begin{equation}
\label{eq:sea_ctrl}
    \tau_\mathrm{cmd} = \tau_s+dk^{-1}\dot \tau_s +bb_\theta^{-1}(\tau_d-\tau_s-d_sk^{-1}\dot \tau_s)
\end{equation}
where $b_\theta$ is the desired reduced motor inertia, $d_s$ is the desired system damping coefficient, and $\tau_d$ is the desired joint torque. Applying (\ref{eq:sea_ctrl}) to (\ref{eq:H_sea}) yields the shaped joint torque dynamics

\begin{equation}
\label{eq:H_sea_cl}
H_{\mathrm{cl}}^{A,i}(s)
=\frac{\tau_a(s)}{\tau_d(s)}
=\frac{1}{b_\theta}
\frac{k+ds}{s^2+d_\theta s+\frac{k}{J_\theta}},
\end{equation}

where $J_\theta$ and $d_\theta$ are the shaped equivalent inertia and damping coefficient, respectively, defined as

\begin{equation}
\label{eq:shaped_values}
J_\theta^{-1}=b_\theta^{-1}+m^{-1},
\qquad
d_\theta=\frac{d_s}{b_\theta}+\frac{d}{m}.
\end{equation}

As for the uncontrolled actuator, the shaped second-order dynamics can be characterized by their natural frequency, damping ratio, and steady-state torque ratio,

\begin{equation}
\label{eq:sea_modal_shaped}
\omega_{n,\theta}=\sqrt{\frac{k}{J_\theta}},
\qquad
\zeta_\theta=\frac{d_\theta}{2}\sqrt{\frac{J_\theta}{k}},
\qquad
\left.\frac{\tau_a}{\tau_d}\right|_{s=0}
=\frac{J_\theta}{b_\theta}.
\end{equation}

These quantities characterize the desired joint-level performance. However, (\ref{eq:H_sea_cl}) assumes ideal motor torque generation and therefore neglects the effects of the inner motor-control dynamics. Including the controlled motor model in (\ref{eq:Hm_simplified}) when applying (\ref{eq:sea_ctrl}) to (\ref{eq:H_sea}) instead yields
\begin{equation}
\begin{split}
&\!H^{A,t}_{cl}(\!s\!)\!\!=\\
&\!\!  \frac{\frac{\omega_{bw}}{b_\theta}(k+ds)e^{\text{-}sT_d}}{\!(\!s\!+\!\omega_{bw}\!)\!(\!s^2\!\!+\!\!\frac{d}{J}s\!+\!\!\frac{k}{J}\!)\!+\!\omega_{bw}e^{\text{-}sT_d}\!\left(\!(\frac{d_s}{b_\theta}-\frac{d}{b})s\!+\!k(\frac{1}{J_\theta}-\frac{1}{J})\!\right)\!}
\end{split}
\end{equation}
which can be simplified using a first order Padé aproximation for the delay $e^{-sT_d}\simeq\frac{1-\frac{T_d}{2}s}{1+\frac{T_d}{2}s}$ to:
\begin{equation}
\begin{split} 
H_{\mathrm{cl}}^{A,c}(s)&=
\frac{\frac{2\omega_{bw}}{T_db_\theta}(k+ds)
\left(1-\frac{T_d}{2}s\right)}
{s^4+a_3s^3+a_2s^2+a_1s+a_0 .}, \\
a_3 &= \frac{2}{T_d}+\omega_{bw}+\frac{d}{J},\\
a_2 &= \frac{2\omega_{bw}}{T_d}+\frac{2d}{JT_d}+\frac{k}{J}
+\omega_{bw}\left(\frac{d}{J}-\frac{d_s}{b_\theta}+\frac{d}{b}\right),\\
a_1 &= \frac{2k}{JT_d}+\frac{2\omega_{bw}d_\theta}{T_d}
+\omega_{bw}k\left(\frac{2}{J}-\frac{1}{J_\theta}\right),\\
a_0 &= \frac{2\omega_{bw}k}{T_dJ_\theta}.
\end{split}
\end{equation}

Tab.~\ref{tab:model_cases} compares the resulting joint torque dynamics under different motor-model assumptions. In each case, the same closed-loop torque transfer function is regrouped according to the additional roots introduced by the motor dynamics. In the ideal case, the joint dynamics of (\ref{eq:H_sea_cl}) are directly recovered. The delay introduces an additional pole $-p$ together with a nearby zero, resulting in an approximately cancelling pole--zero pair. Finite control bandwidth introduces an additional low-pass pole $-\omega_0$, whose location is predominantly determined by $\omega_{bw}$. When both effects are considered, both additional dynamics are present. 

The approximations of $p$ and $\omega_0$ reported in the table directly relate the additional poles to $T_d$ and $\omega_{bw}$, and show how their location affects the resonant frequency and damping of the dominant second-order joint dynamics. The $\mathcal{O}(\cdot)$ terms are used to emphasize the dominant contributions and provide a simplified interpretation of these effects. For a given $T_d$ and $\omega_{bw}$, the exact pole locations, and consequently the exact resonant frequency and damping, can instead be obtained from the roots of the complete characteristic polynomial.This allows $T_d$ and $\omega_{bw}$ to be jointly selected via numerical optimization to satisfy the prescribed joint-level performance.

\section{Classic Discrete Design Requirements}
The classic discrete design is based under the assumption that a first order euler aproximation is sufficient to model the effects of the PMSM under a discretized control, and by extension, a Discretized Continous Controller should be sufficient. Given that, the PI controller is replaced by a tustin based controller as 
\begin{equation}
    C^{c2d}[z]= K^{c2d}_{p,i}\underbrace{(1+\frac{TiT_s}{2})}_{:=\beta}\frac{z-\overbrace{\frac{1-\frac{TiT_s}{2}}{1+\frac{TiT_s}{2}}}^{\alpha}}{z-1}
\end{equation}
and the feedforward terms are kept the same. Given this, we can analyze that the closed loop transfer function of the system would be, considering that the back-emf effects are properly canceled (which are not), it would be:
\begin{equation}
    \begin{split}
        H^d_\mathrm{ff}(z) = \frac{b_\omega r_\omega}{z(z-a_w)-jL\dot\theta_eb_\omega r_\omega}
    \end{split}
\end{equation}
and 
\begin{equation}
\!H^{c2d}_{\mathrm{ff},cl}\!=\!\frac{ K_{p,i}^{c2d}\beta b_{\omega}r_\omega(z-\alpha) } {\!(z\!-\!1)\!\!\left[\!z(z\!-\!a_{\omega}\!)\!-\!jL\dot{\theta}_eb_{\omega}r_\omega\!\right]\!\!\!+\!\!K_{p,i}^{c2d}\beta b_{\omega}r_\omega(\!z\!-\!\alpha\!) \!}
\end{equation}
For the discrete implementation to recover the behavior predicted by the original CT design, several $T_s$-dependent approximations must simultaneously hold:
\begin{equation}
\begin{split}    
r_\omega\approx1+j\dot{\theta}_eT_s,&\qquad a_\omega\approx\left(1-\frac{R}{L}T_s\right)\left(1+j\dot{\theta}_eT_s\right)\\
&b_\omega\approx\frac{T_s}{L}\left(1+j\dot{\theta}_eT_s\right).
\end{split}
\end{equation}
Additionally, the current variation between consecutive samples must remain negligible, such that $\boldsymbol{i}_{dq}[k+1]\approx\boldsymbol{i}_{dq}[k]$, allowing the CT feedforward and cross-coupling compensation to approximate their discrete counterparts:
\begin{equation}
ar_\omega-jLb_\omega\dot{\theta}_e\approx a,\qquad \beta=b^{-1}\frac{T_s}{L}.
\end{equation}
These conditions require a sufficiently small $T_s$ relative to the electrical rotational frequency $\dot{\theta}_e$, the motor dynamics $R/L$, and the current-control bandwidth. Therefore, a high sampling-to-fundamental frequency ratio alone does not guarantee the validity of the CT-based design over the complete operating envelope.

\section{Discrete PMSM Model under Acceleration}

To extend the discrete PMSM model to operation under acceleration, the electrical speed is allowed to vary within each sampling interval. Starting from the stationary-frame dynamics, the exact discrete current evolution is given by
% \begin{strip}
 \begin{equation}
 \label{eq:d_i_alphabeta}
 \begin{split}
\boldsymbol{i_\alphabeta}[k+1] &= \underbrace{e^{-\frac{R}{Lf_s}}}_{:=a}\boldsymbol{i_\alphabeta}[k]+\underbrace{\frac{1-e^{-\frac{R}{Lf_s}}}{R}}_{:=b}\boldsymbol{v_\alphabeta}[k]\\
&-\underbrace{\frac{1}{L}\int_{k}^{(k+1)T_s}e^{-\frac{R}{L}((k+1)T_s-t)}\boldsymbol{\epsilon_\alphabeta}(t)dt}_{:=\boldsymbol{\epsilon_{\alpha\beta}^d}[k]}
\end{split}    
\end{equation}
Assuming constant electrical acceleration over each sampling interval, the electrical angle is expressed as
\begin{equation}
\begin{split}
\theta_e(t)&=\theta_e[kT_s]+(t-kT_s)\dot{\theta}_e[kT_s]\\
&+\frac{(t-kT_s)^2}{2}\ddot{\theta}_e[kT_s],\quad t\in[kT_s,(k+1)T_s]
\end{split}
\end{equation}

Introducing the local time variable $\sigma=t-kT_s$, the discrete back-EMF contribution can then be evaluated as

\begin{equation}
\begin{split}
&\boldsymbol{\epsilon_{\alpha\beta}^{d}}[k]=-\frac{e^{-\frac{R}{L}T_s}}{L}\int_{0}^{T_s}e^{\frac{R}{L}\sigma}\epsilon_{\alpha\beta}(kT_s+\sigma)d\sigma \\
&=-\frac{j k_t e^{-\frac{R}{L}T_s}}{L}\int_{0}^{T_s}\dot{\theta}_e(kT_s+\sigma)e^{\frac{R}{L}\sigma+j\theta_e(kT_s+\sigma)}\,d\sigma \\
%&\boldsymbol{\epsilon_{\alpha\beta}^d}[k]=k_t\frac{e^{-\frac{R}{L}((k+1)T_s}}{L}\int_{k}^{(k+1)T_s}j\dot\theta_e(t)e^{\frac{R}{L}t+j\theta_e(t)}dt\\
&=k_t\frac{e^{-\frac{R}{L}T_s}}{L}\bigg(\int_{0}^{T_s}\left(\frac{R}{L}+j\dot{\theta}_e(kT_s+\sigma)\right)e^{\frac{R}{L}\sigma+j\theta_e(kT_s+\sigma)}\,d\sigma \\
&\quad-\int_{0}^{T_s}\frac{R}{L}e^{\frac{R}{L}\sigma+j\theta_e(kT_s+\sigma)}\,d\sigma\bigg) \\
%&=k_t\frac{e^{-\frac{R}{L}((k+1)T_s}}{L}\times \bigg(\int_{k}^{(k+1)T_s} ( \frac{R}{L}+j\dot\theta_e(t)) e^{\frac{R}{L}t+j\theta_e(t)}dt \\
%&-\int_{k}^{(k+1)T_s}\frac{R}{L}e^{\frac{R}{L}t+j\theta_e(t)}dt\bigg) \\
&=k_t\frac{e^{-\frac{R}{L}T_s}}{L}\bigg(
\left[e^{\frac{R}{L}\sigma+j\theta_e(kT_s+\sigma)}\right]_{0}^{T_s} \\
&\quad-\int_{0}^{T_s}\frac{R}{L}e^{\frac{R}{L}\sigma+j\left(\theta_e[kT_s]+\dot{\theta}_e[kT_s]\sigma+\ddot{\theta}_e[kT_s]\frac{\sigma^2}{2}\right)}\,d\sigma
\bigg)\\
%&=k_t\frac{e^{-\frac{R}{L}((k+1)T_s}}{L}\!\!\times\!\! \bigg( \left[e^{\frac{R}{L}t+j\theta_e(t)}\right]^{(k+1)T_s}_{kT_s} \\
%&-\int_{k}^{(k+1)T_s}\frac{R}{L}e^{\frac{R}{L}t+j\left(\theta_e[kT]+\dot \theta_e[kT]t+\ddot \theta_e[kT]\frac{t^2}{2}\right)}dt\bigg) \\
&=k_t\frac{e^{-\frac{R}{L}T_s}}{L}\bigg(
e^{\frac{R}{L}T_s+j\theta_e[(k+1)T_s]}
-e^{j\theta_e[kT_s]} \\
&\quad-\frac{R}{L}e^{j\theta_e[kT_s]-\frac{\left(\frac{R}{L}+j\dot{\theta}_e[kT_s]\right)^2}{2j\ddot{\theta}_e[kT_s]}}
\int_{0}^{T_s}
e^{\frac{j\ddot{\theta}_e[kT_s]}{2}
\left(\sigma+\frac{\frac{R}{L}+j\dot{\theta}_e[kT_s]}
{j\ddot{\theta}_e[kT_s]}\right)^2}
\!\!d\sigma \!\!\bigg) \\
%&=k_t\frac{e^{-\frac{R}{L}((k+1)T_s}}{L}\times \bigg(e^{\frac{R}{L}(k+1)T_s+j\theta_e[(k+1)T_s]}-e^{\frac{R}{L}kT_s+j\theta_e[kT_s]}\\
%&-\frac{R}{L}e^{j\theta_e[kT]-\frac{\left(\frac{R}{L}+j\dot{\theta}_e[kT]\right)^2}{2j\ddot{\theta}_e[kT]}}\!\!\int_{k}^{(k+1)T_s}e^{\frac{j\ddot{\theta}_e[kT]}{2}\left(t+\frac{\frac{R}{L}+j\dot{\theta}_e[kT]}{j\ddot{\theta}_e[kT]}\right)^2}dt\\
&=\frac{k_t}{L}\bigg(e^{j\theta_e[(k+1)T_s]}-e^{-\frac{R}{L}T_s+j\theta_e[kT_s]}-\frac{R}{L}e^{-\frac{R}{L}T_s+j\theta_e[kT_s]}\times\\
&\frac{e^{-\frac{\left(\frac{R}{L}+j\dot{\theta}_e[kT_s]\right)^2}{2j\ddot{\theta}_e[kT_s]}}}{\sqrt{\frac{j\ddot{\theta}_e[kT_s]}{2}}}\!\!\int_{\xi_0}^{\xi_1}\!e^{\xi^2}\!d\xi\bigg),\!\!\begin{cases}\!\!\xi_0\!\!=\!\!\sqrt{\!\dfrac{j\ddot{\theta}_e[kT_s]}{2}}\dfrac{\frac{R}{L}\!+\!j\dot{\theta}_e[kT_s]}{j\ddot{\theta}_e[kT_s]},\\[6pt]\!\!\xi_1\!\!=\!\!\sqrt{\!\dfrac{j\ddot{\theta}_e[kT_s]}{2}}\!\!\left(\!\!T_s\!+\!\dfrac{\frac{R}{L}\!+\!j\dot{\theta}_e[kT_s]}{j\ddot{\theta}_e[kT_s]}\!\!\right)\!\!\end{cases} \\
%&=\frac{k_t}{L}\bigg(e^{j\theta_e[(k+1)T_s]}-e^{-\frac{R}{L}T_s+j\theta_e[kT_s]} +e^{-\frac{R}{L}((k+1)T_s}\times  \\
%&-\!\!\frac{R}{L}\frac{e^{j\theta_e[kT]-\frac{\left(\frac{R}{L}+j\dot{\theta}_e[kT]\right)^2}{2j\ddot{\theta}_e[kT]}}}{\sqrt{\frac{j\ddot{\theta}_e[kT]}{2}}}\!\!\int_{\sqrt{\frac{j\ddot{\theta}_e[kT]}{2}}\left(k+\frac{\frac{R}{L}+j\dot{\theta}_e[kT]}{j\ddot{\theta}_e[kT]}\right)}^{\sqrt{\frac{j\ddot{\theta}_e[kT]}{2}}\left((k+1)T_s+\frac{\frac{R}{L}+j\dot{\theta}_e[kT]}{j\ddot{\theta}_e[kT]}\right)}\!\!e^{\sigma^2}\!\!d\sigma \\ 
\end{split}    
\end{equation}
The resulting quadratic exponential integral can be expressed in terms of Dawson's integral $D(\cdot)$ through the corresponding change of variables, yielding
\begin{equation}
\begin{split}
&\boldsymbol{\epsilon_{\alpha\beta}^{d}}[k]=\!\frac{k_t}{L}\!\bigg(\!\!e^{j\theta_e[(k+1)T_s]}\!-\!e^{-\frac{R}{L}T_s+j\theta_e[kT_s]}\!-\!\frac{R}{L}e^{-\frac{R}{L}T_s+j\theta_e[kT_s]}\!\\
&\quad\frac{e^{-\frac{\left(\frac{R}{L}+j\dot{\theta}_e[kT_s]\right)^2}{2j\ddot{\theta}_e[kT_s]}}}{\sqrt{\frac{j\ddot{\theta}_e[kT_s]}{2}}}\left[e^{\xi_1^2}D(\xi_1)-e^{\xi_0^2}D(\xi_0)\right]\bigg).\\
&=\frac{k_t}{L}\bigg(e^{j\theta_e[(k+1)T_s]}-e^{-\frac{R}{L}T_s+j\theta_e[kT_s]}-\frac{R}{L}\frac{e^{-\frac{R}{L}T_s+j\theta_e[kT_s]}}{\sqrt{\frac{j\ddot{\theta}_e[kT_s]}{2}}}\times\\
&\quad\left[e^{\frac{R}{L}T_s+j\left(\dot{\theta}_e[kT_s]T_s+\ddot{\theta}_e[kT_s]\frac{T_s^2}{2}\right)}D(\xi_1)-D(\xi_0)\right]\bigg) \\
&=\frac{k_t}{L}\!\bigg[\left(1\!-\!\mathcal{D}_k(T_s)\right)e^{j\theta_e[(k+1)T_s]}\!-\!\left(1\!-\!\mathcal{D}_k(0)\right)e^{-\frac{R}{L}T_s+j\theta_e[kT_s]}\bigg]\\
&\quad \text{where}\\
&\mathcal{D}_k(\sigma)=\frac{\frac{R}{L}}{\sqrt{\frac{j\ddot{\theta}_e[kT_s]}{2}}}D\!\left[\sqrt{\frac{j\ddot{\theta}_e[kT_s]}{2}}\left(\sigma+\frac{\frac{R}{L}+j\dot{\theta}_e[kT_s]}{j\ddot{\theta}_e[kT_s]}\right)\right].\\
\end{split}
\end{equation}
where $\mathcal{D}_k(\sigma)$ contains the contribution associated with the variation of the electrical speed within the sampling interval.
Finally, applying the discrete Park transformation at the updated electrical angle and including the one-sample controller delay yields
\begin{equation}
\begin{split}    
&\boldsymbol{i}_{dq}[k+1]=a_x\boldsymbol{i}_{dq}[k]+b_xr_{k-1}\!\left(\boldsymbol{v}_{dq}[k-1]-\boldsymbol{\epsilon}_{dq}[k]\right),\\
&\boldsymbol{\epsilon}_{dq}[k]=r_{k-1}^{-1}\frac{k_tR}{L}\!\left[\frac{r_k^{-1}\left(1-\mathcal{D}_k(T_s)\right)-a\left(1-\mathcal{D}_k(0)\right)}{1-a}\right],\\
&a_x=ar_k,\qquad b_x=br_k,\qquad r_k=e^{-j\Delta\theta_e[k]},\\
&\Delta\theta_e[k]=\dot{\theta}_e[kT_s]T_s+\ddot{\theta}_e[kT_s]\frac{T_s^2}{2}.\\
\end{split}
\end{equation}
The resulting $(dq)$ model therefore extends the constant-speed formulation by explicitly incorporating the electrical acceleration through $\Delta\theta_e[k]$ and $\mathcal{D}_k(\sigma)$.


\bibliographystyle{IEEEtran}
\bibliography{references}

@article{Gabriel1980FieldOrientedCO,
  title={Field-Oriented Control of a Standard AC Motor Using Microprocessors},
  author={Rupprecht Gabriel and Werner Leonhard and Craig J. Nordby},
  journal={IEEE Transactions on Industry Applications},
  year={1980},
  volume={IA-16},
  pages={186-192},
  url={https://api.semanticscholar.org/CorpusID:14562471}
}

@article{UnifiedPassivitybasedControl2007,
  title = {A {{Unified Passivity-based Control Framework}} for {{Position}}, {{Torque}} and {{Impedance Control}} of {{Flexible Joint Robots}}},
  author = {{Albu-Sch{\"a}ffer}, Alin and Ott, Christian and Hirzinger, Gerd},
  year = 2007,
  month = jan,
  journal = {The International Journal of Robotics Research},
  volume = {26},
  number = {1},
  pages = {23--39},
  issn = {0278-3649, 1741-3176},
  doi = {10.1177/0278364907073776},
  urldate = {2025-03-12},
  copyright = {https://journals.sagepub.com/page/policies/text-and-data-mining-license},
  langid = {english}
}

@article{Spong1987,
    author = {Spong, M. W.},
    title = {Modeling and Control of Elastic Joint Robots},
    journal = {Journal of Dynamic Systems, Measurement, and Control},
    volume = {109},
    number = {4},
    pages = {310-318},
    year = {1987},
    month = {12},
    issn = {0022-0434},
    doi = {10.1115/1.3143860},
    url = {https://doi.org/10.1115/1.3143860},
    eprint = {https://asmedigitalcollection.asme.org/dynamicsystems/article-pdf/109/4/310/5604812/310_1.pdf},
}

@article{Haddadin2024Unified,
  author  = {Haddadin, Sami and Shahriari, Erfan},
  title   = {Unified force-impedance control},
  journal = {The International Journal of Robotics Research},
  year    = {2024},
  volume  = {43},
  number  = {13},
  pages   = {2112--2141},
  doi     = {10.1177/02783649241249194},
  publisher = {SAGE Publications}
}

@article{smidt2021discrete,
  title={Discrete-time sliding mode control based on disturbance observer applied to current control of permanent magnet synchronous motor},
  author={Smidt Gabbi, Thieli and Gr{\"u}ndling, Hilton Ab{\'\i}lio and Padilha Vieira, Rodrigo},
  journal={IET Power Electronics},
  volume={14},
  number={4},
  pages={875--887},
  year={2021},
  publisher={Wiley Online Library}
}

@article{busada2018comments,
  title={Comments on “digital current control in a rotating reference frame—part I: system modeling and the discrete time-domain current controller with improved decoupling capabilities”},
  author={Busada, Claudio Alberto and Jorge, Sebastian Gomez and Solsona, Jorge A},
  journal={IEEE Transactions on Power Electronics},
  volume={34},
  number={3},
  pages={2980--2984},
  year={2018},
  publisher={IEEE}
}

@article{hoffmann2015digital,
  title={Digital current control in a rotating reference frame-Part I: System modeling and the discrete time-domain current controller with improved decoupling capabilities},
  author={Hoffmann, Nils and Fuchs, Friedrich W and Kazmierkowski, Marian P and Schr{\"o}der, Dierk},
  journal={IEEE Transactions on Power Electronics},
  volume={31},
  number={7},
  pages={5290--5305},
  year={2015},
  publisher={IEEE}
}

@ARTICLE{Zhang2021deadbeat,
  author={Zhang, Zhijian and Jing, Long and Wu, Xuezhi and Xu, Wenzheng and Liu, Jingdou and Lyu, Gege and Fan, Zilian},
  journal={IEEE Access}, 
  title={A Deadbeat PI Controller With Modified Feedforward for PMSM Under Low Carrier Ratio}, 
  year={2021},
  volume={9},
  number={},
  pages={63463-63474},
  doi={10.1109/ACCESS.2021.3075486}}

@inproceedings{tiapkinCurrentControllerDesign2020,
  title = {Current {{Controller Design}} of {{Precision Servo Drive}}},
  booktitle = {2020 27th {{International Workshop}} on {{Electric Drives}}: {{MPEI Department}} of {{Electric Drives}} 90th {{Anniversary}} ({{IWED}})},
  author = {Tiapkin, Mikhail and Balkovoi, Aleksandr and Samygina, Elizaveta},
  year = 2020,
  month = jan,
  pages = {1--6},
  publisher = {IEEE},
  address = {Moscow, Russia},
  doi = {10.1109/IWED48848.2020.9069577},
  urldate = {2025-03-05},
  copyright = {https://ieeexplore.ieee.org/Xplorehelp/downloads/license-information/IEEE.html},
  isbn = {978-1-7281-4158-9},
  langid = {english}
}

@article{blaschkePrincipleFieldOrientation1972,
  author  = {Blaschke, Felix},
  title   = {The Principle of Field Orientation as Applied to the New Transvector Closed-Loop Control System for Rotating-Field Machines},
  journal = {Siemens Review},
  volume  = {34},
  number  = {3},
  pages   = {217--220},
  year    = {1972}
}

@article{gabrielFieldOrientedControl1980,
  author  = {Gabriel, Rupprecht and Leonhard, Werner and Nordby, Craig J.},
  title   = {Field-Oriented Control of a Standard {AC} Motor Using Microprocessors},
  journal = {IEEE Transactions on Industry Applications},
  volume  = {IA-16},
  number  = {2},
  pages   = {186--192},
  year    = {1980},
  month   = {mar},
  doi     = {10.1109/TIA.1980.4503770}
}

@article{hematiRobustNonlinearControl1990,
  author  = {Hemati, N. and Thorp, J. S. and Leu, M.-C.},
  title   = {Robust Nonlinear Control of Brushless {DC} Motors for Direct-Drive Robotic Applications},
  journal = {IEEE Transactions on Industrial Electronics},
  volume  = {37},
  number  = {6},
  pages   = {460--468},
  year    = {1990},
  month   = {dec},
  doi     = {10.1109/41.103449}
}

@article{kimDiscreteTimeCurrentRegulator2010,
  title = {Discrete-{{Time Current Regulator Design}} for {{AC Machine Drives}}},
  author = {{Hongrae Kim} and Degner, Michael W and Guerrero, Juan M and Briz, Fernando and Lorenz, Robert D},
  year = 2010,
  month = jul,
  journal = {IEEE Transactions on Industry Applications},
  volume = {46},
  number = {4},
  pages = {1425--1435},
  issn = {0093-9994, 1939-9367},
  doi = {10.1109/TIA.2010.2049628},
  urldate = {2025-03-05},
  copyright = {https://ieeexplore.ieee.org/Xplorehelp/downloads/license-information/IEEE.html},
  langid = {english}
}

@article{rowanNewSynchronousCurrent1986,
  author  = {Rowan, Timothy M. and Kerkman, Russel J.},
  title   = {A New Synchronous Current Regulator and an Analysis of Current-Regulated {PWM} Inverters},
  journal = {IEEE Transactions on Industry Applications},
  volume  = {IA-22},
  number  = {4},
  pages   = {678--690},
  year    = {1986},
  month   = {jul},
  doi     = {10.1109/TIA.1986.4504778}
}

@article{brizAnalysisDesignCurrent2000,
  author  = {Briz, Fernando and Degner, Michael W. and Lorenz, Robert D.},
  title   = {Analysis and Design of Current Regulators Using Complex Vectors},
  journal = {IEEE Transactions on Industry Applications},
  volume  = {36},
  number  = {3},
  pages   = {817--825},
  year    = {2000},
  month   = {may},
  doi     = {10.1109/28.845057}
}

@article{holtzDesignFastRobust2004,
  author  = {Holtz, Joachim and Quan, Juntao and Pontt, Jorge and Rodr{\'i}guez, Jos{\'e} and Newman, Patricio and Miranda, Hern{\'a}n},
  title   = {Design of Fast and Robust Current Regulators for High-Power Drives Based on Complex State Variables},
  journal = {IEEE Transactions on Industry Applications},
  volume  = {40},
  number  = {5},
  pages   = {1388--1397},
  year    = {2004},
  month   = {sep},
  doi     = {10.1109/TIA.2004.834049}
}

@article{mohamedDesignImplementationRobust2007,
  title = {Design and {{Implementation}} of a {{Robust Current-Control Scheme}} for a {{PMSM Vector Drive With}} a {{Simple Adaptive Disturbance Observer}}},
  author = {Mohamed, Y.A.-R.I.},
  year = 2007,
  month = aug,
  journal = {IEEE Transactions on Industrial Electronics},
  volume = {54},
  number = {4},
  pages = {1981--1988},
  issn = {0278-0046},
  doi = {10.1109/TIE.2007.895074},
  urldate = {2025-03-05},
  copyright = {https://ieeexplore.ieee.org/Xplorehelp/downloads/license-information/IEEE.html},
  langid = {english}
}

@article{Liao2017deadbeat,
author = {Liao, Yong and Li, Fu and Lin, Hao and Zhang, Jimiao},
title = {Discrete current control with improved disturbance rejection for surface-mounted permanent magnet synchronous machine at high speed},
journal = {IET Electric Power Applications},
volume = {11},
number = {7},
pages = {1333-1340},
doi = {https://doi.org/10.1049/iet-epa.2017.0005},
url = {https://ietresearch.onlinelibrary.wiley.com/doi/abs/10.1049/iet-epa.2017.0005},
eprint = {https://ietresearch.onlinelibrary.wiley.com/doi/pdf/10.1049/iet-epa.2017.0005},
year = {2017}
}

@article{walzDahlinBasedFast2019,
  author  = {Walz, Stefan and Lazar, Radu and Buticchi, Giampaolo and Liserre, Marco},
  title   = {Dahlin-Based Fast and Robust Current Control of a {PMSM} in Case of Low Carrier Ratio},
  journal = {IEEE Access},
  volume  = {7},
  pages   = {102199--102208},
  year    = {2019},
  doi     = {10.1109/ACCESS.2019.2927402}
}

@article{busadaSynchronousReferenceFrame2020,
  author  = {Busada, Claudio A. and G{\'o}mez Jorge, Sebasti{\'a}n and Solsona, Jorge A.},
  title   = {A Synchronous Reference Frame {PI} Current Controller With Dead Beat Response},
  journal = {IEEE Transactions on Power Electronics},
  volume  = {35},
  number  = {3},
  pages   = {3097--3105},
  year    = {2020},
  month   = {mar},
  doi     = {10.1109/TPEL.2019.2925705}
}

@inproceedings{iskandarJointLevelControl2020,
  author    = {Iskandar, Maged and Ott, Christian and Eiberger, Oliver and Keppler, Manuel and Albu-Sch{\"a}ffer, Alin and Dietrich, Alexander},
  title     = {Joint-Level Control of the {DLR} Lightweight Robot {SARA}},
  booktitle = {2020 IEEE/RSJ International Conference on Intelligent Robots and Systems (IROS)},
  pages     = {8903--8910},
  year      = {2020},
  month     = {oct},
  doi       = {10.1109/IROS45743.2020.9340700}
}

@article{Yim2009Modified,
  author  = {Jung-Sik Yim and Seung-Ki Sul and Bon-Ho Bae and Nitin R. Patel and Silva Hiti},
  title   = {Modified Current Control Schemes for High-Performance Permanent-Magnet {AC} Drives With Low Sampling to Operating Frequency Ratio},
  journal = {IEEE Transactions on Industry Applications},
  year    = {2009},
  volume  = {45},
  number  = {2},
  pages   = {763--771},
  month   = {mar},
  doi     = {10.1109/TIA.2009.2013600}
}

\newpage

\vfill

\end{document}